\documentclass[conference]{IEEEtran}
\IEEEoverridecommandlockouts
\usepackage{cite}
\usepackage{amsmath,amssymb,amsfonts}
\usepackage{algorithmic}
\usepackage{graphicx}
\usepackage{textcomp}
\usepackage{xcolor}

\usepackage{booktabs}
\usepackage{amsthm}
\usepackage{multirow}
\usepackage{multicol}
\usepackage{adjustbox}
\usepackage{kotex}

\def\BibTeX{{\rm B\kern-.05em{\sc i\kern-.025em b}\kern-.08em
    T\kern-.1667em\lower.7ex\hbox{E}\kern-.125emX}}
\begin{document}

\newcommand\nj[1]{\textcolor{black}{#1}} 
\newcommand\knj[1]{\textcolor{black}{#1}} 
\newcommand\jin[1]{\textcolor{black}{#1}}
\newcommand\sh[1]{\textcolor{black}{#1}}
\newcommand\shtwo[1]{\textcolor{black}{#1}}
\newcommand\js[1]{\textcolor{black}{#1}}
\newcommand\kjs[1]{\textcolor{black}{#1}}
\newcommand\jh[1]{\textcolor{black}{#1}}
\newcommand\kjh[1]{\textcolor{black}{#1}}
\newcommand\je[1]{\textcolor{black}{#1}}

\title{Pose estimator and tracker \\using temporal flow maps for limbs\\
{\footnotesize 
}
\thanks{* Corresponding author}
\thanks{This work was supported by the ICT R\&D program of MSIP/IITP, Korean Government (2017-0-00306).}
}

\author{\IEEEauthorblockN{Jihye Hwang\textsuperscript{1}, Jieun Lee\textsuperscript{2}, Sungheon Park\textsuperscript{1} and Nojun Kwak\textsuperscript{1*}}
\IEEEauthorblockA{\textit{\textsuperscript{1}Department of Transdisciplinary Studies, Seoul National University, Korea} \\
\textit{\textsuperscript{2}Department of Electrical and Computer Engineering, Ajou University, Korea}\\
hjh881120@snu.ac.kr, mokona8585@gmail.com, 
sungheonpark@snu.ac.kr, nojunk@snu.ac.kr}
}

\maketitle

\begin{abstract}

For human pose estimation in videos, it is significant how to use temporal information between frames. \jh{In this paper, we propose temporal \jin{flow} maps \jh{for limbs (TML)} and a multi-stride method to estimate and track human poses.} \jh{The proposed temporal \jin{flow} maps \jin{are} unit vectors describing the limbs' movements.}
We constructed a network to learn both spatial information and temporal information end-to-end. Spatial information such as joint heatmaps and part affinity fields \jin{is} regressed in the spatial network part, and 
\jh{the TML} is regressed in the temporal network part. We also propose \shtwo{a data augmentation method
}to learn various types of 
\jh{TML }better. The proposed multi-stride method expands the data by randomly selecting two frames within a defined range. We demonstrate that the proposed method efficiently estimates and tracks human poses on the PoseTrack 2017 and 2018 datasets.

\end{abstract}


\section{Introduction}

Human pose estimation (HPE) is \nj{one} of the most significant tasks in  \knj{computer} vision.
\nj{Over the past few years, static image-based pose estimation for either a single person or multiple \sh{people} has achieved high accuracy using convolutional neural networks (CNNs).}
\nj{Deeply-structured networks as well as iterative networks have been proposed for this task \shtwo{to take advantage of} their large receptive fields and \shtwo{rich} representation power.}

\jh{In case of multiple people pose estimation, there \shtwo{are} two major approaches: top-down and bottom-up approaches.}
The bottom-up approach \cite{cao2017realtime, doering2018joint, insafutdinov2016deepercut, Iqbal_CVPR2017, jin2017towards, fractalnet, deepcut16cvpr, MPR, DBLP:conf/cvpr/XiaWCY17, Zhu2017} detects the body joints of all \sh{people} \knj{at once} and then estimates human poses individually. 
On the other hand, the top-down approach \cite{fang2017rmpe, girdhar2018detecttrack, DBLP:conf/nips/NewellHD17, jin2017towards, George2017, xiao2018simple, xiu2018pose} consists of a human detector that detects human bounding boxes and a single person pose estimator that locates and groups body joints in each bounding \knj{box}. 

Recently, \sh{HPE in videos has grabbed \knj{attentions} as an extension of HPE in a single image.} For HPE in videos, \sh{human pose tracking should be performed as well as the pose estimation.}  Many researches have \shtwo{exploited} temporal information in various ways for tracking. 
\knj{Such methods as a bounding box tracking algorithm, optical flow, similarity of estimated shape, temporal flow fields (TFF) and so on have been applied for this task} \cite{xiao2018simple, MPR, xiu2018pose, doering2018joint}.

\nj{Among them, the work of Xiao et al.} \cite{xiao2018simple} is a representative top-down approach. They detected the pose based on the extracted bounding  boxes \nj{and} proposed a box tracking method \sh{for pose tracking,} \nj{which} is \sh{a combination of} a box propagation using optical flow and a flow-based pose similarity. 
Likewise, \je{Ibrahim et al.} \cite{MPR} used the similarity of the \shtwo{poses} to track the pose \shtwo{while} \sh{it} is \sh{a bottom-up approach.} 
In case of \cite{xiu2018pose}, they proposed \nj{an} online pose tracking \nj{algorithm} called \nj{\textit{pose flow}, which is } 
an association of \nj{the} same person in different frames. They created the optimized pose flow using several scores such as mean score of all keypoints. 
In \shtwo{summary}, these studies proposed tracking methods based on the similarity of estimated poses.  

\begin{figure*}[t!]
\centering
  \includegraphics[width=18cm]{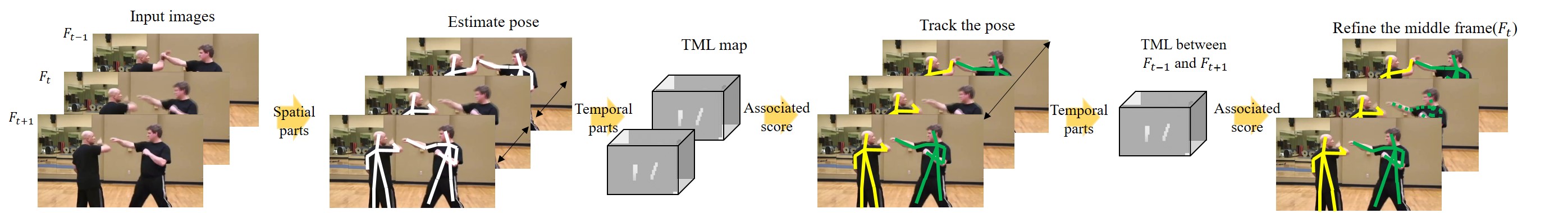}
  \caption{An inference flow of proposed method. A set of frames ($F_{t-1}$, $F_{t}$, $F_{t+1}$) is defined for temporal inference. Two frames that are ($F_{t-1}$, $F_{t}$) or ($F_{t}$, $F_{t+1}$) are input into the network as a pair.
  First, poses are estimated using the spatial part on the each frame and 
  \jh{TML} are extracted by the temporal part at the same time. To track poses, we calculate the association score of each person using the 
  \jh{TML }and the joint distance score. The optimal connection is found by using a bipartite method. 
  In order to refine the middle of frame $F_t$, we need to get the associated information between ($F_{t-1}$, $F_{t+1}$) through the 
  \jh{TML }and the joint distance. If the pose is connected between $F_{t-1}$ and $F_{t+1}$, we added the average pose of between ($F_{t-1}$, $F_{t+1}$).  
  Note that, the time interval of 
  \jh{TML }is 1 \nj{at inference stage but it can be a greater number at training stage which will be described in the multi-stride method.}}
  \label{fig:inference}
\end{figure*}

On the other \nj{hand}, \je{Andreas et al.} \cite{doering2018joint} represented an association of \shtwo{poses} as \shtwo{temporal vector maps} \nj{called \textit{temporal flow fields}} (TFF). TFF indicates \nj{the flow of a joint} between two frames. They estimated poses through the heatmaps and \textit{part affinity fields} \cite{cao2017realtime} and \nj{used} a similarity measure in a bipartite graph matching to track the poses.
However, when \sh{estimating} \nj{TFF}, using only joint location \sh{may} not \jh{be} enough to track the poses. \sh{Tracking only a single joint may \nj{lead to a} lack of representation power or may \nj{be} vulnerable to occlusion of \nj{joints}. Therefore, if a limb that connects two joints is tracked, it is \nj{expected} to provide richer representation for the tracker and to enhance robustness to occlusion. In addition, considering frames of multiple strides rather than only two consecutive frames can further improve the robustness and performance of the network.} 

\sh{To this end,} \nj{in} this paper, we propose a \nj{pose estimator and tracker based on 
\jh{TML} which is designed to} \je{represent a temporal movement of \nj{a} person \nj{\shtwo{by estimating} the direction of limbs' movement.} 
\nj{More specifically, we} subdivide \nj{each limb into several \je{sections equally} \je{in each frame}}. \sh{Then, 2D unit vector\je{s} that represent the direction of corresponding limb \je{sections} between two frames \je{are} calculated, which \je{are} used to build \je{each limb's} temporal map\je{s}.}} 
\je{
\nj{A huge amount of data} is needed to train the \jh{TML} because \shtwo{the maps have to learn} extensive information. 
Thus, we \shtwo{develop} a multi-stride method as a data augmentation \nj{method} to learn various \shtwo{types of} \jh{TML}. \sh{In other words}, we randomly take the two frames within \nj{a} given time \nj{range}.}

\nj{Figure \ref{fig:inference} shows the overall flow of inference in the proposed method.} 
During inference, we \jh{process three frames as a frame set at a time. }
First, we extract poses on each frame \js{in the form of joint heatmaps }and part affinity fields at the spatial part. 
\jh{The extracted poses are tracked between the first and the second frame of the three frames by associated scores obtained from a TML score and a joint distance.}
The second frame lies in the middle of the consecutive three frames. 
\js{After the same procedure is applied to the second frame and the third frame, the second frame is refined by analyzing the association scores of the three frames. 
\jh{The frame set is selected at one frame interval. }
This makes the model stable since the information from the frames back and forth adjust the result of the intermediate frame.}

\jh{Thus, the contributions of our work are as follows:}

1) We propose \jh{the TML}
to represent directions of \nj{limbs' movement.} 

2) A multi-stride method is proposed to train \nj{various} 
\jh{TML} as a data augmentation method. 

3) \jh{The current poses are refined by associated score of the previous frame and the next frame.}
    
We evaluated the proposed method on the PoseTrack 2017 and 2018 \nj{datasets} \cite{PoseTrack}. To prove the effectiveness of the proposed method, we \nj{made a comparison of our work with state-of-the-art algorithms}.

\begin{figure*}
\centering
  \includegraphics[width=17cm]{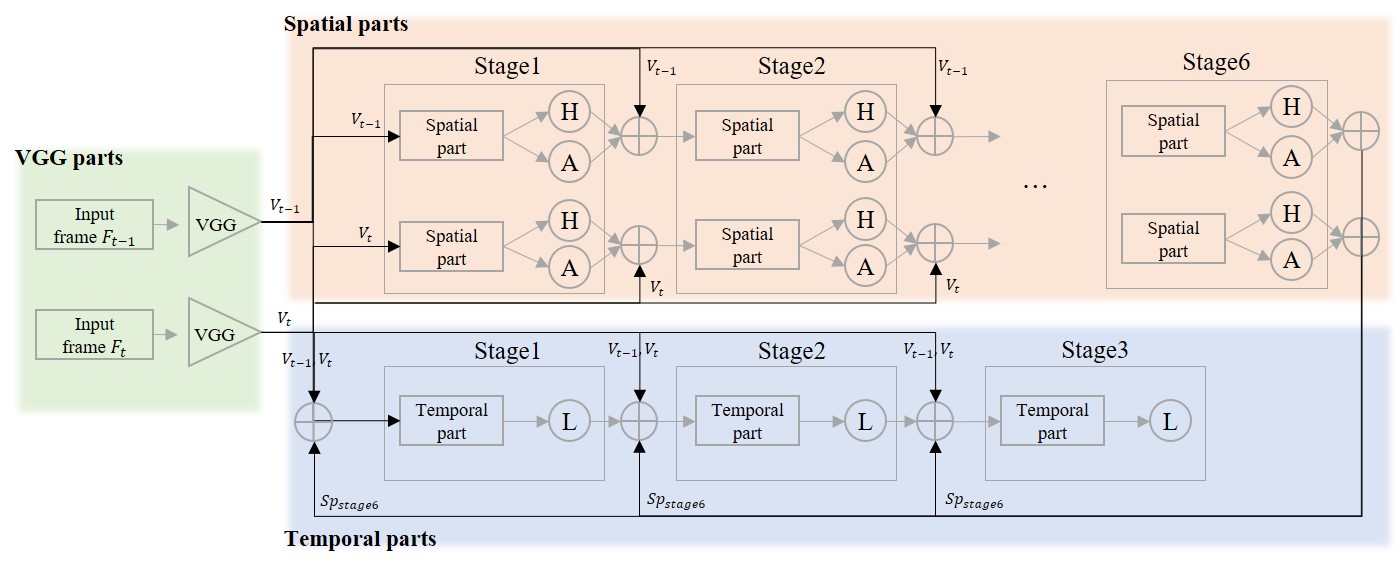}
  \caption{The structure of \nj{the} proposed network. \nj{The spatial and the temporal parts} are combined \sh{together in a single network}. On the spatial \sh{part}, \sh{joint} heatmaps (H circle) and \sh{part} affinity fields (A circle) are regressed. \sh{Outputs from the spatial part} and \sh{features from the final layers of the VGG parts} are fed into the \sh{temporal} part. \nj{The temporal part} regresses 
  \jh{the TML }(L circle). \sh{Pixel-wise L2 losses are used to optimize all the outputs.} \jh{$V_{t-1}$ and $V_t$ mean the extracted features from VGG parts at $t-1$ and $t$ frame respectively. $Sp_{stage6}$ is the concatenated spatial features of $t-1$ frame and $t$ frame at the stage6.}}
  \label{fig:structure}
\end{figure*}


\section{Related work}



\subsection{Single person pose estimation}
\je{Over the past few years, m}any \je{CNN based} methods \js{in \je{single person} pose estimation \cite{bulat2016human, ICCV2017Chen, DBLP:conf/cvpr/ChuYOMYW17, insafutdinov2016deepercut, newell2016stacked, deepcut16cvpr, posemachines2014, DBLP:conf/cvpr/TompsonGJLB15, Tompson:2014:JTC:2968826.2969027, cpm2016, DBLP:conf/iccv/YangLOLW17} \je{used}} very deep networks.
\js{Also, a recursive methodology has been adopted in many competitive methods.}  

Newell et al. \cite{newell2016stacked} \je{proposed \kjs{a model with} multiple hourglass modules that} repeat\kjs{s} bottom-up \nj{and} top-down processing and 
\kjs{Wei et al. \cite{cpm2016} proposed a convolution version of the pose machine \cite{posemachines2014} which has been proposed by Ramakrishna et al.}
\js{Features in these} networks \js{possess} a large receptive field \js{which extracts an efficient representation of \nj{human} context. }
\js{Advances in the} single person pose estimation \js{have made it possible to proceed research on \nj{multi-person pose} estimation.}

\subsection{Multi person pose estimation}
Multi-person pose estimation \cite{cao2017realtime, doering2018joint, fang2017rmpe, girdhar2018detecttrack, insafutdinov17arttrack,insafutdinov2016deepercut, Iqbal_CVPR2017,  jin2017towards, DBLP:conf/nips/NewellHD17, fractalnet, George2017, deepcut16cvpr, MPR, DBLP:conf/cvpr/XiaWCY17, xiao2018simple, xiu2018pose, Zhu2017} \nj{methods can be categorized as} \shtwo{top-down and bottom-up} approaches. 
\je{The t}op-down approach\shtwo{es} \cite{fang2017rmpe, girdhar2018detecttrack, DBLP:conf/nips/NewellHD17, jin2017towards, George2017, xiao2018simple, xiu2018pose} firstly detect a \nj{person's} bounding box and \nj{estimate} \shtwo{single} pose on the extracted bounding box.
On the other hand, \nj{the bottom-up approach\shtwo{es} \cite{cao2017realtime, doering2018joint, insafutdinov2016deepercut, Iqbal_CVPR2017, jin2017towards, fractalnet, deepcut16cvpr, MPR, DBLP:conf/cvpr/XiaWCY17,Zhu2017} firstly detect parts of people and then determine} \shtwo{poses in the input image by connecting the candidate parts.}

Cao et al. \cite{cao2017realtime} \nj{proposed} part affinity fields (PAFs) to associate body parts and \nj{determined} the pose using the PAFs.
\kjs{Doering et al. \cite{doering2018joint} and Zhu et al. \cite{Zhu2017} suggested modified version of methods based on this model.} DeeperCut \cite{deepcut16cvpr} is a graph decomposition method to re-define a variable number of consistent body part configurations.
\nj{The performances of state-of-the-art multi-person pose estimation methods are pretty good for a single frame. However, to apply the methods on real applications, we need to combine them with tracking algorithms for video data.}


\subsection{Human pose estimation with tracking}
Several methods have been proposed to estimate and track \nj{human poses on videos} \cite{doering2018joint, girdhar2018detecttrack, insafutdinov17arttrack, Iqbal_CVPR2017, jin2017towards, xiao2018simple, xiu2018pose}. \nj{These methods can be divided into two groups depending on whether \kjs{the} learned temporal information is used or not.}
\shtwo{For the methods that do not use the} learned temporal information, they track the pose by applying optical flow, box tracking algorithm, and so on. Xiu et al. \cite{xiu2018pose} proposed a pose tracker based on a pose flow that is a flow structure indicating the same person in different frames by pose distance. Xiao et al. \cite{xiao2018simple} tracked the pose to use a flow-based pose tracking algorithm based on box propagation using optical flow and a flow-based pose similarity. 
Instead of \kjs{naively} connecting the relationships between detected poses, several papers trained sequential information. Radwan et al. \cite{MPR} used a bi-directional long-short term memory (LSTM) framework to learn the consistencies of the human body shapes. Doering et al. \cite{doering2018joint} proposed temporal flow fields that \je{are} vector fields to indicate the direction of joints. 

\nj{However, tracking only a single joint may not contain enough temporal information due to lack of representation power and also it may be vulnerable to occlusion of joints. Therefore, in this paper, instead of a single joint point, a limb connecting two adjacent joints is tracked, which is expected to resolve the above mentioned problems. Also, \sh{during both training and testing}, we consider a pair of frames with more than one time interval rather than only using two consecutive frames  for the robustness of the proposed architecture.}



\section{Method}
\label{sec:method}

%

\je{In order to estimate and track human poses \shtwo{using} a single network, we constructed a network \nj{consisting} of \nj{two sub-parts (a spatial part and a temporal part)} as shown in Figure \ref{fig:structure}. We used the network \nj{presented in} \cite{cao2017realtime} for \kjs{the} spatial part, which has iterative \sh{stages}. The stage \nj{consists} of two branches, \sh{one} for \sh{joint} heatmaps and \sh{the other for part} affinity \sh{fields}. In the proposed network, we take two frames \sh{as inputs}. \jh{Each frame passes through the VGG network to \shtwo{extract features} \cite{simonyan2014very}. The features \shtwo{of each frame} are fed into \shtwo{each spatial part of the network} in parallel. The features \shtwo{of two input frames} and the output of the spatial part's last stage are concatenated and fed into the temporal part.} 
The temporal part has \nj{a single} branch to train \jh{the TML. }
Same as \nj{the} spatial part, we apply the iterative stages. Since \sh{the last stage outputs of the spatial part are fed into each stage} of the temporal part, the spatial and the temporal information affect each other through the end-to-end learning.} We calculate the pixel-wise L2 loss \nj{as a loss function} \jh{for each map} at all stages.

\nj{Below is a more detailed description on each part.} 
\begin{itemize} 
\item Spatial part\jh{s}: \jh{Spatial part}\jin{s are} made up \nj{of} six stages to learn the \sh{joint} heatmaps and \sh{the part affinity fields}. VGG features of two frames are fed into each spatial part. The losses of \sh{the joint} heatmaps (\(H\) circle \shtwo{in Figure \ref{fig:structure}}) and \sh{the part affinity fields} (\(A\) circle \shtwo{in Figure \ref{fig:structure}}) are calculated \nj{at each stage as in} \cite{cao2017realtime}.
 
\item Temporal part\jh{s}: The temporal parts \shtwo{resemble the single branch of the spatial part and} \jin{are} made up of three stages to learn 
\jh{the TML}. \shtwo{Each stage of the temporal part}\jin{s} has \nj{three} $3\times3$ convolutions and \nj{two} $1\times1$ convolutions. 
 \nj{The first} stage takes the concatenated features \nj{as inputs}: the VGG features \nj{of \sh{the two input} frames}, \sh{and the joint} heatmaps and \sh{the part affinity fields from}  the last stage \nj{in} the spatial part\jin{s}. \nj{The second and the third stages} additionally \nj{use the \jh{TML} 
  of the previous stage as an input}.
 Iterative stages \sh{gradually improve the accuracy of} \jh{the TML} 
 The loss of \jh{the TML }
 (\(L\) circle \shtwo{in Figure \ref{fig:structure}}) is calculated \nj{at each stage} using a pixel-wise L2 loss function. \nj{The number of stages, three, was found experimentally.} 
\end{itemize} 

\nj{As the spatial parts are the same network presented in \cite{cao2017realtime}, we will focus on the temporal part in the following subsections. }


\begin{figure}[t!]
    \centering
    \includegraphics[width=\linewidth]{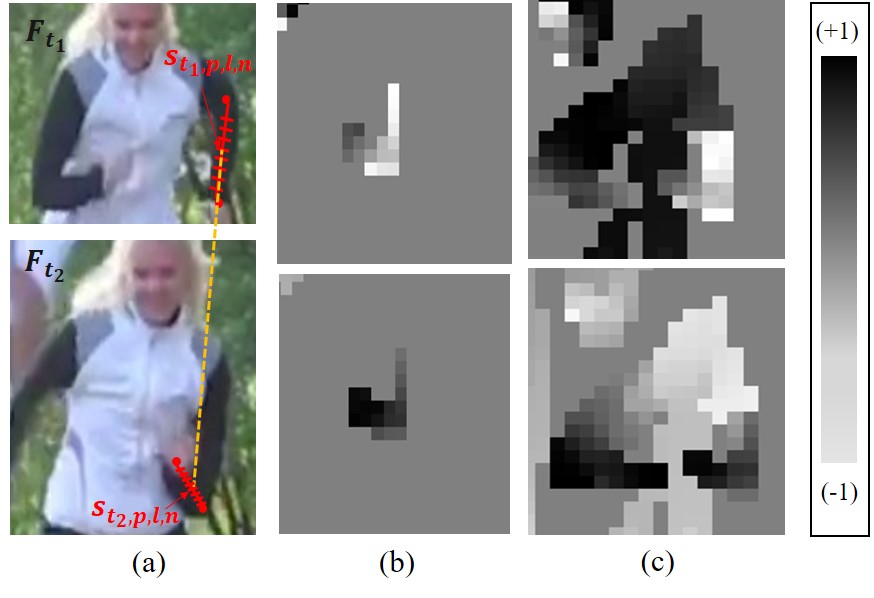}
    \caption{\nj{An example of} \jh{TML} 
    (a) \sh{Illustration explaining} \nj{how to obtain the \jh{TML}
    using the frames $F_{t_1}$ and $F_{t_2}$}. We \nj{subdivide each limb} into several parts and \nj{calculate} \je{the} unit vector of each pair (connected by the yellow lines, \(S_{t_1,p,l,n}\) and \(S_{t_2,p,l,n}\)). (b) Visualization of the left arm \jh{TML} 
    on \sh{$x$(top) and $y$(bottom) coordinates.} (c) \je{Accumulated  TML 
    for all limbs on \sh{$x$(top) and $y$(bottom) coordinates}. \sh{The values of \jh{TML}
    are between the range of -1 and 1}.}}
    \label{fig:fv}
\end{figure}
        
    

\subsection{Temporal flow Maps for Limb movement (TML)}

\jh{The TML }is \kjs{a set of} \nj{vector field\je{s}} representing flow\sh{s} of \je{person's limbs. \kjs{In this paper, \shtwo{a} limb denotes a \jh{part} 
linking two joints \jh{such as an upper arm and calf}.} \sh{An example of the} vector fields of a single limb is} shown in Figure \ref{fig:fv}(b). To \nj{obtain \je{these} new type} \sh{of} map\je{s}, first we divide \nj{each limb} at regular intervals to \nj{multiple} parts. 
Figure \ref{fig:fv}(a) shows \nj{an} example of \nj{a} divided limb \nj{at} frame \(F_{t_{1}}\) and \(F_{t_{2}}\). \nj{In the figure, the red} line means \nj{a limb 
\jh{and each yellow dot line shows}} \kjs{the same} parts \jh{on the same limb} between frames. \nj{A separate\shtwo{d} part} ($S_{t,p,l,n}$), \shtwo{which means} \nj{an $n$-th separated part on $l$-th limb on the $p$-th person at the frame $t$, is} used to calculate \nj{the movement} direction between two frames. 
Based on the pair $(S_{t_1,p,l,n}, S_{t_2,p,l,n})$, we calculate a unit vector $v$ \nj{as follows: }
\begin{equation}
v = \frac{({S}_{t_1,p,l,n} - {S}_{t_2,p,l,n})}{\left \| {S}_{t_1,p,l,n}-{S}_{t_2,p,l,n} \right \|_{2}}.
\label{equ:unitvector}
\end{equation}
\nj{Here, $n$, $l$ and $p$ represent the index of a separated part, a limb and a \nj{person} respectively, and $t_1$ and $t_2$ are the  frame indices. The part $S$ is represented by a two dimensional vector corresponding to the position of the part and thus $v$ is also a two-dimensional vector.}

Then, the \(L\) \nj{for the $l$-th limb} is encoded through \nj{the} unit vector \(v\) for each pixel $s=(x,y)$ \jh{which is the limb passes through at the time interval $t_1$ and $t_2$}.
\jh{To draw the TML, we applied the similar process of part affinity field in \cite{cao2017realtime}. }

\begin{equation}
    L_{l,p}(s) = \left\{\begin{matrix} 
\begin{matrix}
v & \textrm{if }~ s \subseteq C \\ 
0 & \textrm{otherwise}.
\end{matrix}
\end{matrix}\right.
\label{equ:limbflow}
\end{equation}

\jh{According to the condition ($C$), each pixel is determined to whether it is on the path of limb movement at the time interval $t_1$ and $t_2$.}
\nj{More concretely, in our case, the pixels belonging to the line segment ($S_{t_1,p,l,n}$, $S_{t_2,p,l,n}$) with a constant width is filled with the value of $v$ and the other pixels remain as zero.}

\sh{When the \jh{TML}
of multi person are overlapped at the same position, it is averaged to preserve the scales of the output. Thus,} the final \jh{TML}
\sh{for the $l$-th joint} averages the \jh{TML}
of \sh{the joints of all people appeared in the image as follows:} 
\begin{equation}
     L_{l}(s) = \begin{cases}
          \frac{1}{P(s)}\sum_{p=1}^{P(s)} L_{l,p}(s), & \quad \textrm{if } P(s) \ge 1 \\
          0 & \quad \textrm{if } P(s) = 0,
          \end{cases}
\end{equation}
where $P(s)$ means the number of non-zero vectors at pixel $s$. \kjh{All of $n$ divided parts follow the above process to make the \jh{TML.}}
Figure \ref{fig:fv}(b) shows \nj{a} \sh{visualization of } \jh{the TML }
of \sh{a single limb that is a part of left \jh{lower }arm in} $x$ and $y$ \sh{directions}. The closer \nj{the value to $-1 (+1)$, the brighter (darker) it becomes.} 
\nj{In Figure \ref{fig:fv}(a), we can see that the left hand of the person moves to the} left-down side. Then, the direction of $x$ channel is $-$ \nj{while} the direction of $y$ channel \nj{becomes} $+$ as shown in Figure \ref{fig:fv}(b). Thus, it \shtwo{is} confirmed that the direction is different in each pixel \nj{of the \jh{TML.}}


Unlike optical flow \cite{horn1981determining} \nj{representing} \shtwo{directions and magnitudes} at each location, the \jh{TML}
only represents the direction\shtwo{s} using unit vector\shtwo{s}. \nj{Because the \jh{TML}
does not contain magnitude information, it is more prone to change of time interval between frames}.
\shtwo{The multi-stride method for data augmentation, which will be describe in the next subsection, helps to alleviate this issue and successfully trains the network using video frames with different sampling rates.}

Furthermore, \nj{the} \jh{TML}
channel \nj{can be set as an individual channel for each limb \sh{(Figure \ref{fig:fv}(b))} or as an accumulated channel which accumulates \sh{ 
\jh{the TML}
of all limbs (Figure \ref{fig:fv}(c))}. } The \nj{number of individual channels becomes} \textit{ the number of limbs \(\times\) 2} ($x, y$ coordinate channel) \shtwo{while} the accumulated channel has \shtwo{only} 2 channels ($x, y$ coordinate channel). We \nj{will show} the efficiency \nj{of different types of channel in the} evaluation section.

\subsection{Multi-stride method}

\jh{The TML} has temporal information on the joint location. We need \nj{a} huge dataset \nj{containing} various situations and poses to learn the map\shtwo{s}. If \shtwo{one trains} \jh{the TML }by using only \shtwo{a fixed} time interval in video\shtwo{s} sequentially, \shtwo{only limited types of maps can be obtained, which usually contains very small movements.} Thus, \nj{we propose a multi-stride method which uses a pair of frames with various time interval as a data augmentation method}. To generate the various time interval, we randomly select two frames within a given time range.

\begin{figure}[t!]
\centering
  \includegraphics[width=\linewidth]{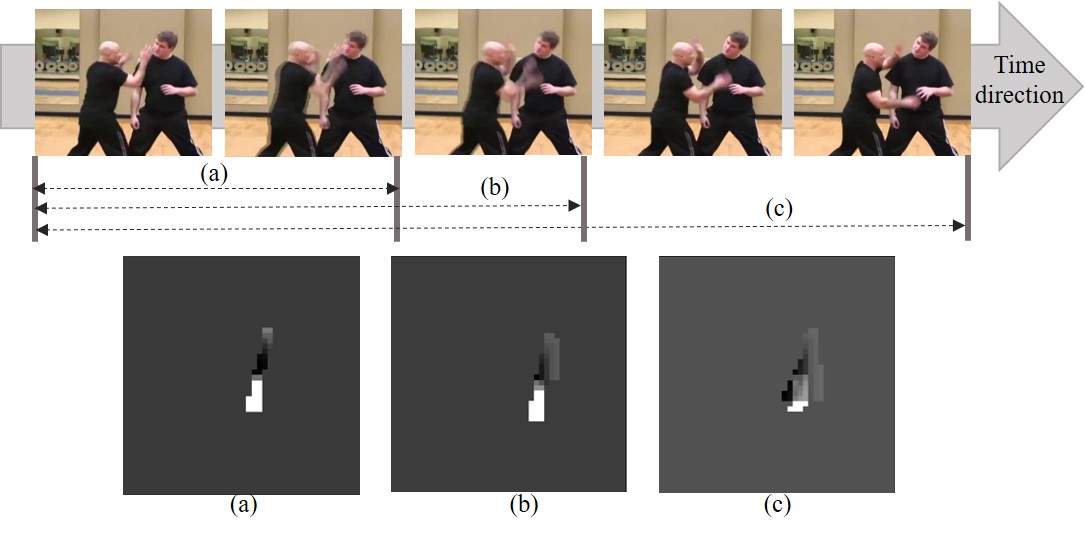}
  \caption{\nj{Examples} of 
  \jh{the TML} \je{of x coordinate} with various time intervals. \sh{Consecutive image sequences are shown from left to right}. (a), (b) and (c) are the \je{right arm} 
  \jh{TML }of \je{the left person with} the different time intervals\je{, 1, 2 and 4 respectively. Using various strides, it is possible to get 
  \jh{the TML }of both small and large movements.} 
}
  \label{fig:multi-stride}
\end{figure}

Figure \ref{fig:multi-stride} shows the examples of 
\jh{ the TML }on \nj{a} small-motion video. If \nj{we} only use the time interval \nj{of one shown in} Figure \ref{fig:multi-stride}(a), we cannot get a direction of large motions in this video. On the other hands, if \nj{we} use various time intervals, additional maps can be obtained as shown in Figure \ref{fig:multi-stride}(b) and (c), 
\nj{making it possible} to express a case where the motion is varied even in a small movement.

Furthermore, our multi-stride \nj{method} can be used to refine pose\shtwo{s} at inference time. The proposed \nj{refining} method \nj{can be useful when a frame misses a person but the preceding and the next frames successfully target the person.}
\nj{In this case, 
because our multi-stride method randomly selected two frames at the training time and the network learned this situation}, we can extract 
\jh{the TML }and track the pose between frame $F_t$ and $F_{t+2}$. More \nj{details can be found} in the Section \ref{sec:inference}.

\begin{table*}[t!]
\centering
  \caption{The estimation and tracking results of \nj{the} proposed methods on the PoseTrack2017 and 2018 validation data. \nj{\#stage} means the number \nj{o\je{f }stacked stages in the} temporal part. Joint-Flow has \nj{a} different type of temporal map that is created by joint movement. Basically, \nj{the} proposed 
  \jh{TML }\je{has} \nj{two channels ($x$ and $y$)} for each limb. ($*$) means \sh{the method in which 
  \jh{the TML}
  of all limbs are accumulated in a single map for x and y directions.}
  \(+\) adopted the non-maximum suppression (NMS) for joints. \(++\) \nj{indicates that the proposed refining method for the middle frame pose is applied}. Distance means that \nj{\jh{The TML}
  is not used in the calculation of the association score in (\ref{equ:score}) by setting $\alpha$ to 0, which means that it only uses the torso distance of a person for the associated score.} }
    \begin{adjustbox}{width=0.8\textwidth}
      \centering

    \begin{tabular}{c|c|c|c|c|c|c|c|c|c|c|c}
    \toprule
    data  & \multicolumn{2}{c|}{Method} & \multicolumn{8}{c|}{MOTA}                                     & mAP \\
          &  & \#stage & Head  & Shou  & Elb   & Wri   & Hip   & Knee  & Ankl  & Total &  \\
    \midrule
    \midrule
          & Joint-Flow($*$) & 1     & 70.6  & 70.1  & 50.6  & 37.5  & 53.9  & 41.8  & 30.3  & 52    & 73.1 \\
          & Joint-Flow & 1     & 48.5  & 48.3  & 30.3  & 19.3  & 33.9  & 23    & 13.5  & 32.1  & 73.2 \\
          & \jh{TML}($*$) & 1     & 72    & 70.6  & 52.1  & 37.7  & 53.8  & 41.3  & 30.9  & 52.6  & 71.3 \\
    2017  & \jh{TML} & 1     & 70.1  & 69.5  & 51.9  & 40.5  & 53.8  & 43.5  & 32.7  & 52.9  & 72.9 \\
\cmidrule{2-12}          & \jh{TML} & 3     & 74.7  & 74.1  & 61.7  & 49.4  & 59    & 52.6  & 43.7  & 60.3  & 70.9 \\
          & \jh{TML}+ & 3     & 75.1  & 74.6  & 62.5  & 50.1  & 59.5  & 53    & 44.2  & 60.9  & 71.3 \\
          & Distance+ & 3     & 49.9  & 50.1  & 40.5  & 31.5  & 37.7  & 32.4  & 26.7  & 39.2  & 71.3 \\
          & \jh{TML}++ & 3     & 75.5  & 75.1  & 62.9  & 50.7  & 60    & 53.4  & 44.5  & 61.3  & 71.5 \\
    \midrule
    2018  & \jh{TML}++ & 3     & 76    & 76.9  & 66.1  & 56.4  & 65.1  & 61.6  & 52.4  & 65.7  & 74.6 \\
    \bottomrule
    \end{tabular}%
    \end{adjustbox}
  \label{tab:val_2017}%
\end{table*}%

\subsection{Inference}
\label{sec:inference}


We define \nj{a set of frames consisting} of three frames ($F_{t-1}$, $F_{t}$, $F_{t+1}$) as shown in Figure \ref{fig:inference} \nj{for temporal inference which associates joint candidates in different frames}. 
First, on \nj{each} frame, we estimate the joint candidates using the \nj{joint} heatmaps and \nj{spatially} connect the candidates using part affinity fields as in \cite{cao2017realtime}. The heatmaps and part affinity fields are created by \nj{the} spatial part \nj{as shown in Figure \ref{fig:structure}}.

Based on the connected joint candidates \shtwo{denoted as $I$}, we track the poses. 
We calculate the associated score of each person in different frames. The associated score  \nj{is calculated by a linear combination of} a score of \jh{the TML}
(\(S_{\jh{T}}\)) and a score of joint distance (\(S_{d}\)): 
\begin{equation}
S = \alpha S_{\jh{T}}+(1-\alpha) S_{d},
\label{equ:score}
\end{equation}
\nj{where $\alpha$ is a hyper-parameter which is set to 0.5 in our experiments.}

We measure the score of a candidate movement on each \jh{TML}
by calculating the line integral. More specifically, we extract two joint candidates $I_j^{t_1}$ and  $I_j^{t_2}$ in different frames at time $t_1$ and $t_2$ corresponding to the joint $j$ and make a normalized directional vector between the two joint candidates. Then the value of the \jh{TML}
corresponding to the line segment ($I_j^{t_1}$, $I_j^{t_2}$) is obtained to take inner product with the directional vector. This is done for all the points in the line segment and integrated as follows:
\begin{equation}
    S_{\jh{T}} = \frac{1}{n_J}\sum_{j=1}^{n_J}\int_{u=0}^{u=1}L_{l}(K(u))\cdot \frac{I_{j}^{t_1}-I_{j}^{t_2}}{\left \| I_{j}^{t_1}-I_{j}^{t_2} \right \|_{2}} du.
\end{equation}
Here, $I$ is a joint candidate and $n_J$ is the number of \nj{joints for a person which is determined in the spatial part}. 
$K(u)$ indicates interpolated points in the line segment ($I_j^{t_1}$, $I_j^{t_2}$) where $u \in \{0,1\}$, i.e., $K(u)=(1-u)\cdot I_j^{t_1}+u\cdot I_j^{t_2}$.
This score measures the plausibility of joint association between frames using the 
\jh{TML}. 

We measured the joint distance (\(S_{d}\)) between the frames using \nj{the Euclidean} distance. 
\begin{equation}
    S_{\jh{d}} = \frac{1}{n_J}\sum_{j=1}^{n_J}\begin{Vmatrix}
I_{j}^{t_1} - I_{j}^{t_2} \end{Vmatrix}
\end{equation}
Both scores are \nj{given a different weight by using the variable \(\alpha\) which} is determined through experiments. Finally, we find the optimal connection by applying a bipartite graph \cite{doering2018joint}.

\nj{After this,} we refine the poses that \nj{have disappeared} at the intermediate frame and come out again as shown in \nj{the second and the third images in} Figure \ref{fig:inference}. More specifically, those situation means that the pose are not extracted on the \nj{frame} $F_{t}$ but extracted and tracked between the \nj{frames} $(F_{t-1}, F_{t+1})$.
To improve the situation, after \nj{ pairs of frames with the time interval of one  ($F_{t-1}$, $F_{t}$), ($F_{t}$, $F_{t+1}$) are processed, then to recover the missing person or joint in the frame $F_t$, a pair of frames with the time interval of two ($F_{t-1}$, $F_{t+1}$) is inputted to the proposed network followed by the above tracking method.
After that, \shtwo{the poses or the joints missed in the middle of the frame $F_{t}$ are filled with} average location\shtwo{s of those in} $F_{t-1}$ and $F_{t+1}$. 
}

\section{Experiments}

\subsection{Datasets}

In order to prove \nj{the efficiency} of the proposed method, experiments on the PoseTrack 2017 and 2018 datasets \cite{PoseTrack} \nj{have been} performed. PoseTrack datasets are large-scale benchmarks for human pose estimation and tracking. The PoseTrack datasets have \sh{various videos of human activities including} fishing, running, tennis and so on. The datasets \sh{include a wide range of pose variations} from a monotonous pose to a complex pose.
PoseTrack datasets have the videos more than 500 sequences that are expected \nj{to be} more than 20K frames. It is \nj{composed} of 250 videos for \nj{training}, 50 videos for validation and 214 videos for \nj{testing}. PoseTrack 2018 dataset annotated more data than 2017. 

The annotation types of PoseTrack 2017 and 2018 are different. The joints of PoseTrack 2018 added \nj{more} parts such as ears and shoulder on \nj{top of} the joints of PoseTrack 2017 and a new  \nj{order of joints} has been set. 
However, at the test time, mean average precision (mAP), multiple object tracker accuracy (MOTA) and multiple object tracking precision (MOTP) are evaluated in the annotation order of PoseTrack 2017. 

\subsection{Implementation details}
We used the open-source library Caffe \cite{jia2014caffe} to \nj{implement} our model. Our model was trained with a weight decay of 0.0005, a momentum \nj{of} 0.9 and \nj{a} learning rate \nj{of} 0.00005. We used the pre-trained model \sh{of \cite{cao2017realtime} trained} on COCO \sh{keypoints dataset} \cite{COCO} as \sh{a} base network. At the \nj{training time}, we \nj{needed} to change the joint order and to add some parts such as ears to middle of head from Postrack to COCO to use \nj{the} pre-trained model parameter, because COCO data and PoseTrack data have different order of \nj{joints}. 

At the \nj{training time}, we applied \nj{data augmentation methods} such as random crop, random rotate and so on. 
We set the scaling and rotation \nj{parameters} based on the first frame among the two images. After a scaling and a rotation, we randomly select a person and crop a region \nj{such that the center of the selected person \shtwo{is located at} the center of the region}. The scaling, the rotation and selected person information of the first frame are applied equally in the second frame. 



\subsection{Evaluation}

MOTA, MOTP and mAP are used to evaluate the performance \cite{milan2016mot16}. Table \ref{tab:val_2017} shows the results of \nj{the} proposed methods by different settings - \nj{using different numbers (1 or 3) of} iterative stages \nj{in} the temporal part \nj{(\#stage)}, using channel accumulation of 
\jh{TML }instead of using individual channels for each joint ($*$), and a tracking method only using distance score \nj{by setting $\alpha$ in (\ref{equ:score}) as 0} (distance). \kjh{Through the experiment, we empirically decide the number of subdivide each limb to 20 pieces to make the 
\jh{TML.}}

To \sh{make the temporal network part having as few parameters as possible while maintaining} high performance, we experimented with \sh{different} number of repetition stages, 1, 3 and 6. 
The spatial part used a fixed six stages. \nj{T}able \ref{tab:val_2017} only \nj{compares the performances with one and three  iterative stages in the temporal part}, because the \nj{experimental} result of the iterative 6 stages is lower than \nj{that of 3 stages} and has a huge \nj{number of} parameters. 

Similar \nj{to optical} flow \cite{horn1981determining}, we accumulate all limb movements in one map called accumulated channel map as shown in Figure \ref{fig:fv}(c). On the Table \ref{tab:val_2017}, ($*$) means that the network used the accumulated \jh{TML}. Basically, we use a map with a channel for each limb called individual channel map. The number of channel on individual channel map is (the number of $(x,y)$ channels = 2)$\times$(the number of limbs), but the accumulated channel map has \nj{only two $(x,y)$ channels}.
In all \nj{the tested} networks, the accumulated channel map obtained lower accuracy than \nj{the} \shtwo{individual channel} map. \shtwo{Huge amount of the} \nj{directional information} of each limb is lost in the accumulated map, because the map includes some problems, \nj{e.g., different limbs overlap in the same location and have an averaging effect on that point.} 

\begin{table}[t!]
  \centering

    \caption{\sh{Pose estimation and tracking performance} on PoseTrack 2017 test dataset.}
  \begin{adjustbox}{width=1\linewidth}

  \centering
     \begin{tabular}{r|c|c|c|c|c|c}
    \toprule
    \multicolumn{2}{c|}{Method} & mAP   & MOTA  & MOTP  & Prec. & Rec \\
    \midrule
    \midrule
    \multicolumn{1}{c|}{\multirow{3}[6]{*}{Top-down}} & Poseflow\cite{xiu2018pose} & 63    & 51    & 16.9  & 71.2  & 78.9 \\
\cmidrule{2-7}          & MVIG  & 63.2  & 50.8  & -     & -     & - \\
\cmidrule{2-7}          & Xiao et al.\cite{xiao2018simple} & 74.6  & 57.8  & 62.6  & 79.4  & 80.3 \\
    \midrule
          & JointFlow\cite{doering2018joint} & 63.6  & 53    & 23.2  & 82.1  & 70.6 \\
\cmidrule{2-7}    \multicolumn{1}{p{5.125em}|}{Bottom-up} & Jin et al.\cite{jin2017towards} & 59.16 & 50.59 & -     & -     & - \\
\cmidrule{2-7}          & TML++  & 68.78 & 54.46 & 85.2  & 80    & 76.1 \\
    \bottomrule
    \end{tabular}%
    \end{adjustbox}

  \label{tab:comp_other}%
\end{table}%

\begin{table}[t!]
  \centering
  \caption{\sh{Pose estimation and tracking performance} on PoseTrack 2018 test dataset.}
    \begin{adjustbox}{width=1\linewidth}

    \begin{tabular}[width=\linewidth]{c|c|c|c|c|c}
    \toprule
    Method & Additional training data & MOTA  & mAP   & Wrists AP & Ankles AP \\
    \midrule
    \midrule
    Xiao et al. \cite{xiao2018simple} & +COCO+Other      & 61.37 & 74.03 & 73    & 69.05 \\
    \midrule
    ALG   & +COCO+Other      & 60.79 & 74.85 & 72.62 & 71.11 \\
    \midrule
    Miracle & +COCO+Other      & 57.36 & 70.9  & 68.19 & 66.06 \\
    \midrule
    CMP   & +COCO       & 54.47 & 64.67 & 61.78 & 60.86 \\
    \midrule
    PR    & +COCO       & 44.54 & 59.05 & 50.16 & 49.4 \\
    \midrule
    TML++  & +COCO       & 54.86 & 67.81 & 60.2  & 56.85 \\
    \bottomrule
    \end{tabular}%
    \end{adjustbox}
  \label{tab:comp_other2018}%
\end{table}%

\begin{figure*}[t!]
\centering
  \includegraphics[width=\linewidth]{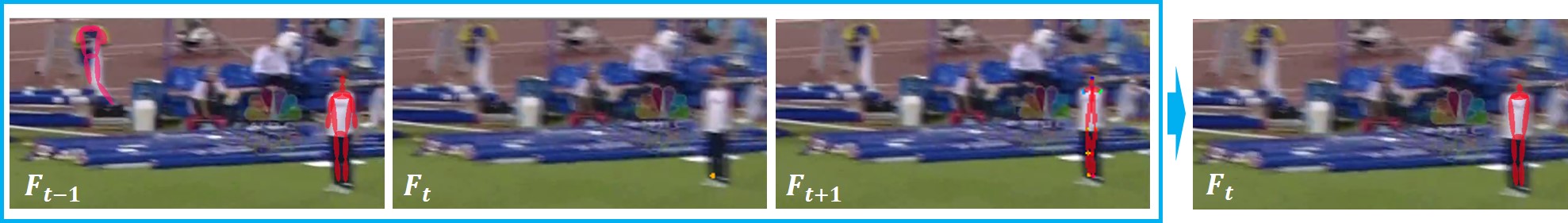}
  \caption{\sh{An example of pose refinement using multi-stride inputs during the inference}. The person at the right side \sh{of input images} (red line) is tracked from $F_{t-1}$ to $F_{t+1}$, but \sh{the pose of the person is not detected} at $F_t$. \sh{By associating the poses at $F_{t-1}$ and $F_{t+1}$, we can retrieve the missed pose at $F_t$}. On the other hand, we cannot refine the person on the left side (pink line), because it is only estimated at the $F_{t-1}$.
 }
  \label{fig:refine}
\end{figure*}

\begin{figure*}[t!]
\centering
  \includegraphics[width=\linewidth]{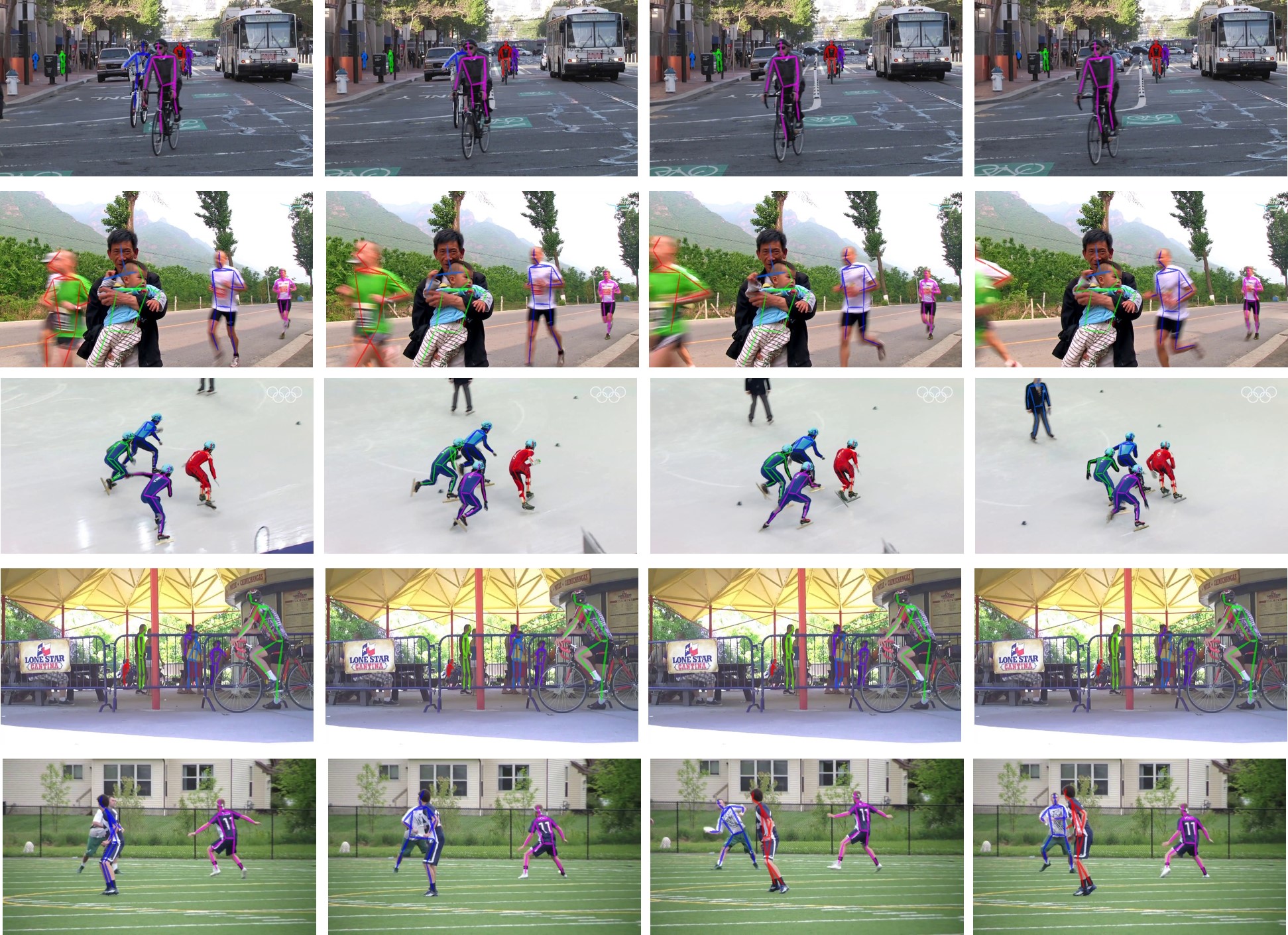}
  \caption{The \shtwo{qualitative} results of the proposed multi-stride pose estimator and tracker. The images are in chronological order from left to right. \sh{Tracked poses are displayed in the same color.}}
  \label{fig:tracking}
\end{figure*}

We \shtwo{implemented} and compared \shtwo{the performance of the} Joint-Flow map to show that \shtwo{the} map created using limb\shtwo{s} is more efficient than \shtwo{the} map created using joint\shtwo{s}. The Joint-Flow map is constructed as a direction in which the joint moves between two frames. The Joint-Flow map \nj{follows} the equation of \nj{(\ref{equ:limbflow}) but uses} the joint location instead of separated part $s$.
The mAPs of Joint-Flow are higher than 
\jh{the TML}
, but MOTAs are lower. 
This results shows the difficulty of tracking using the Joint-Flow, because the Joint-Flow map has less information than the 
\jh{TML}
. Moreover, we compared with JointFlow \cite{doering2018joint} that proposed a temporal map about joint movement as shown in Table \ref{tab:comp_other}. On the PoseTrack 2017 test set, our results \nj{are better than those of} the JointFlow \cite{doering2018joint}.

Because the proposed method is the bottom-up approach, it is possible to detect \nj{many} joint candidates on the same part. 
Thus, a non-maximum suppression (NMS) is applied for joints to reduce \nj{confusion} after estimating joint location. \shtwo{(+) in Table \ref{tab:val_2017}} means that first we detect joints using the \nj{joint} heatmaps and refine the joint using NMS. Reducing the confusing candidates increases tracking performance \nj{by around 0.4\% in mAP and 0.6\% in MOTA}.

The sum of the 
\jh{TML}
score and the joint distance score is used for the association score to track poses. We experimented \nj{to see} how the joint distance affects to association score. On \nj{T}able \ref{tab:val_2017}, (Distance) means that only \nj{joint distances of a person is used in the calculation of association score}. \nj{To enable this, at inference time}, we use the same structure as 
\jh{TML+}
and \nj{set} \(\alpha\) to \(0\). 
Only using the distance score \nj{incurs more confusion} with nearby people \nj{and the resultant MOTA is by far} lower than others on average. However, we need to use the distance score to handle the case of no motion. Thus, we apply \(\alpha\) to \(0.5\) \nj{in all the other cases}.

One of our \nj{contributions} is the refining method for the middle frame pose. We refine the pose on the middle of frame by analyzing between three frames. \shtwo{(++) in Table \ref{tab:val_2017},  Table \ref{tab:comp_other}, and  Table \ref{tab:comp_other2018}} means that \nj{the refining method is applied}. 
Figure \ref{fig:refine} shows \nj{an} example result of the refined pose. The pose on \nj{$F_t$} is refined through the association \nj{between frames $F_{t-1}$ and $F_{t+1}$}. In case of \nj{the} person on the right side (red line), the person is tracked at the $F_{t-1}$ and the $F_{t+1}$, but not tracked at the $F_t$. Through the \nj{refining} method, an average pose between $F_{t-1}$ and $F_{t+1}$ \nj{is} added \nj{on the frame} $F_t$. 
Unfortunately, the person on the left side (pink line) can not be tracked through the \nj{refining} method, because the pose is not estimated at the $F_{t+1}$.  

Figure \ref{fig:tracking} shows \shtwo{qualitative results of} pose \nj{estimation} and tracking. Poses are estimated and tracked well in a variety \nj{of} environments \nj{even when} several people move close together or quickly. 
Because our association score \nj{considers} the distance score, poses that have a little movement can \nj{also} be tracked as shown in the fourth row on Figure \ref{fig:tracking}.  
Unfortunately, if the poses nearly \nj{occludes} each other as in the last row of Figure \ref{fig:tracking}, the pose is likely \nj{to be} missed. For future work, \nj{we may propagate the pose} through the 
\jh{TML }
and \nj{refine} the estimated pose \nj{by} comparing \nj{it} with the propagated pose \shtwo{to address this}.

We compare \nj{our method with the} state-of-the-art \nj{methods} on the PoseTrack 2017 and 2018 test datasets as shown in Table \ref{tab:comp_other}. Though the proposed method \nj{shows a lower performance} than the highest record \cite{xiao2018simple}, the result of \nj{the} proposed network is the best \nj{among} the bottom-up approaches. 

Because the PoseTrack challenge was held on the September 2018, papers using the PoseTrack 2018 data have not been published yet. We could \nj{not} compare \nj{the proposed method} with other methods on the PoseTrack 2018 validate data. However, we can compare results of state-of-the-art on the PoseTrack 2018 test data through the results on the PoseTrack leader-board site as shown in the Table \ref{tab:comp_other2018}. We cannot \shtwo{compare the structures of the networks}, but ours shows the best performance among the \nj{ones} \nj{trained only using} COCO data.

\section{Conclusions}

We propose a multi-stride pose estimator and tracker. It tracks the joints based on the \jh{TML}
 which is \nj{a unit vector map} representing the human flow. 
The multi-stride method has been used to train various temporal flow maps. 
Our method utilizes both the spatial and temporal information.
Spatial information such as joint heatmaps and part affinity fields is regressed by the spatial part and 
\jh{TML }is regressed by the temporal part. 
The combined network can be trained in an end-to-end manner influencing each other.
We demonstrate the efficiency of the proposed method on the PoseTrack 2017 and 2018 datasets.





{\small

\bibliographystyle{IEEEtran}
\bibliography{conference_041818}
}

\end{document}